\documentclass[conference]{IEEEtran}

\usepackage{times}
\usepackage{epsfig}
\usepackage{graphicx}
\usepackage{amsmath}
\usepackage{amssymb}
\graphicspath{{figures/}} 

\usepackage{amsmath,amsfonts,bm}

\def\eqref#1{equation~\ref{#1}}

\def\1{\bm{1}}

\def\mA{{\bm{A}}}

\def\mE{{\bm{E}}}

\def\mM{{\bm{M}}}

\DeclareMathAlphabet{\mathsfit}{\encodingdefault}{\sfdefault}{m}{sl}
\SetMathAlphabet{\mathsfit}{bold}{\encodingdefault}{\sfdefault}{bx}{n}
\newcommand{\tens}[1]{\bm{\mathsfit{#1}}}
\def\tA{{\tens{A}}}

\def\tC{{\tens{C}}}
\def\tD{{\tens{D}}}

\def\tK{{\tens{K}}}

\def\tP{{\tens{P}}}

\def\tR{{\tens{R}}}

\def\tT{{\tens{T}}}
\def\tU{{\tens{U}}}
\def\tV{{\tens{V}}}

\def\tX{{\tens{X}}}

\def\tZ{{\tens{Z}}}

\usepackage{multirow}
\usepackage{arydshln}
\usepackage[ruled,vlined,linesnumbered]{algorithm2e}

\usepackage{url}
\usepackage{comment}
\usepackage[dvipsnames]{xcolor}

\newcommand{\bestvid}[1]{\textcolor{BrickRed}{#1}}

\newcommand{\bestste}[1]{\textcolor{NavyBlue}{#1}} %

\def\etal{{\it et~al.}}
\def\eg{{\it e.g.}}
\def\ie{{\it i.e.}}

\begin{document}

\title{Improved Point Transformation Methods For Self-Supervised Depth Prediction}

\author{
\IEEEauthorblockN{Chen Ziwen}
\IEEEauthorblockA{Grinnell College\\
Grinnell, IA\\
{\tt\small chenziwe@grinnell.edu}}
\and
\IEEEauthorblockN{Zixuan Guo}
\IEEEauthorblockA{Grinnell College\\
Grinnell, IA\\
{\tt\small guozixua@grinnell.edu}}

\and
\IEEEauthorblockN{Jerod Weinman}
\IEEEauthorblockA{Grinnell College\\
Grinnell, IA\\
{\tt\small jerod@acm.org}}
}

\maketitle

\begin{abstract}
   Given stereo or egomotion image pairs, a popular and successful method for unsupervised learning of monocular depth estimation is to measure the quality of image reconstructions resulting from the learned depth predictions. 
   Continued research has improved the overall approach in recent years, yet the common framework still suffers from several important limitations, particularly when dealing with points occluded after transformation to a novel viewpoint.
   While prior work has addressed this problem heuristically, this paper introduces a z-buffering algorithm that correctly and efficiently handles occluded points. 
   Because our algorithm is implemented with operators typical of machine learning libraries, it can be incorporated into any existing unsupervised depth learning framework with automatic support for differentiation.
   Additionally, because points having negative depth after transformation often signify erroneously shallow depth predictions, we introduce a loss function to penalize this undesirable behavior explicitly. 
   Experimental results on the KITTI data set show that the z-buffer and negative depth loss both improve the performance of a state of the art depth-prediction network.
   The code is available at \url{https://github.com/arthurhero/ZbuffDepth}. 
\end{abstract}


\section{Introduction}\label{sec:intro}

\begin{figure}
\begin{center}
{\includegraphics[width=0.98\linewidth]{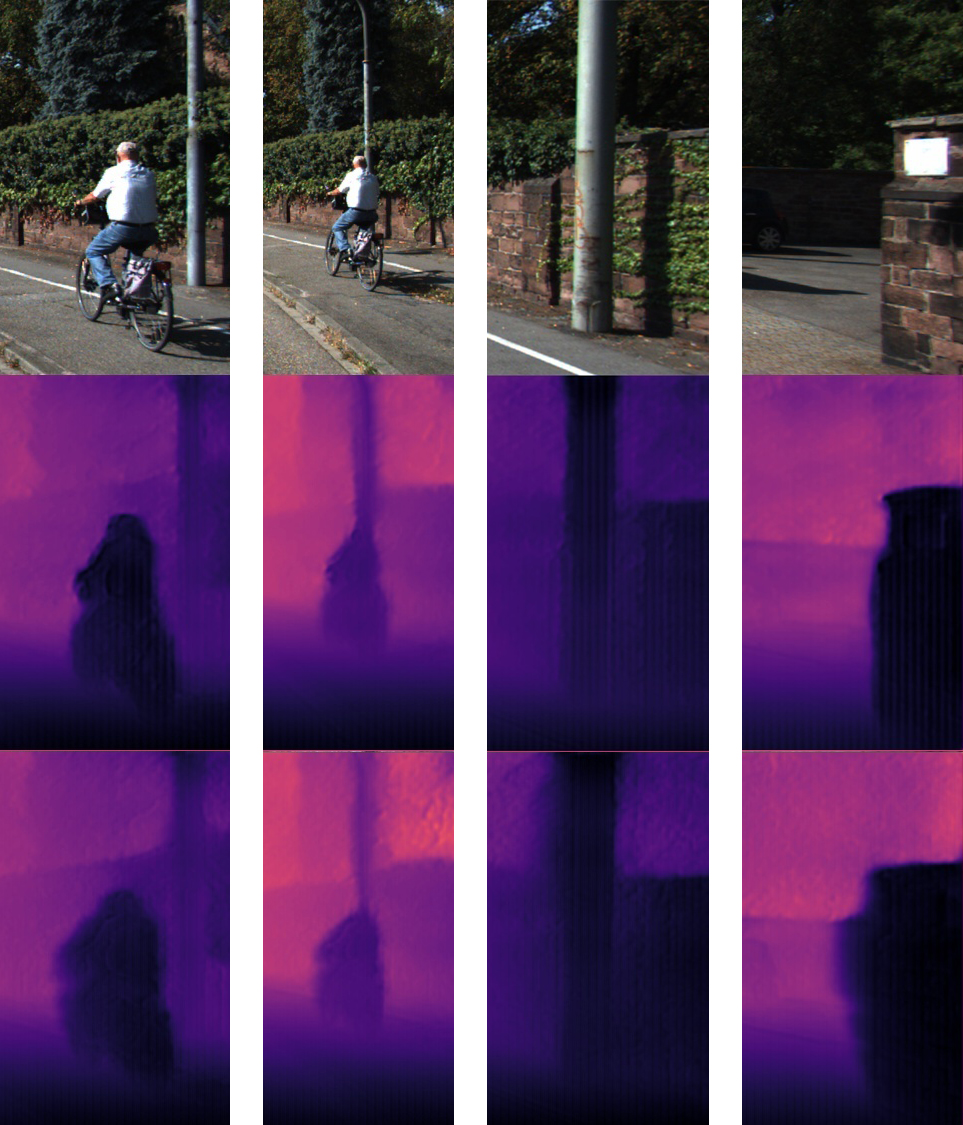}}
\end{center}
   \caption{Comparison of depth prediction results with (second row) and without (third row) z-buffer in the self-supervised training pipeline.
     Differences are most pronounced at large depth boundaries where occlusions are most prominent.
     (Images from the KITTI~\cite{kitti} Eigen test split.)
   }
\label{fig:demo}
\end{figure}

With the advent of deep neural networks and large, real-world data sets, it has become possible to learn models that estimate depth from a single image with remarkable accuracy.
The common framework for learning to predict depth from a single image involves training on image pairs of a scene with known relative camera positions, either from stereo or egomotion. 
Models can be trained in a self-supervised fashion (meaning there is no ground truth depth signal) by using one image and the predicted depth to reconstruct the view in the other image.
The training should bring the reconstructed image and the true image into agreement.

\begin{figure*}
\begin{center}
{\includegraphics[bb=90bp 150bp 700bp 370bp,clip,width=0.8\linewidth]{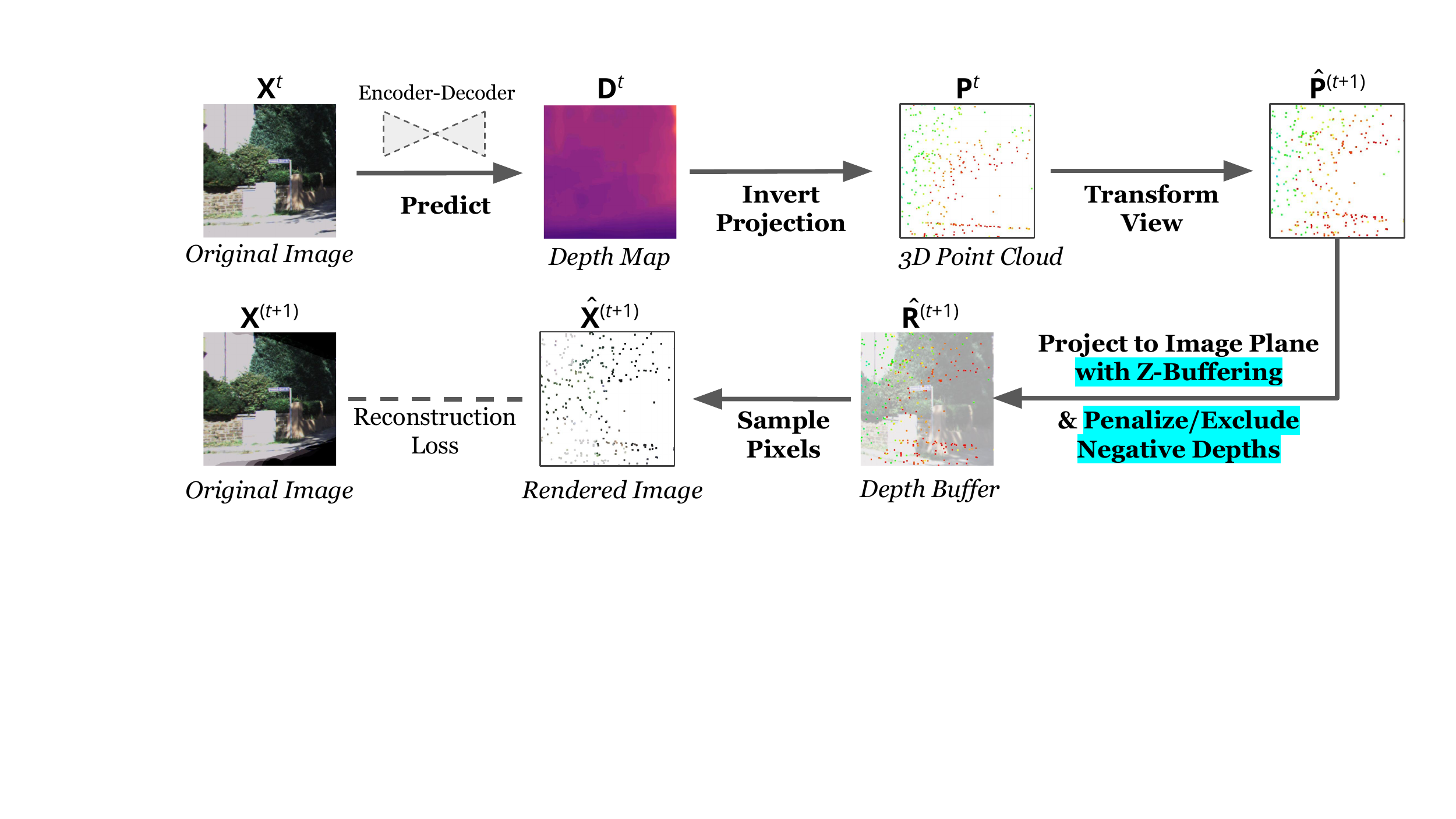}}
\end{center}
   \caption{Overview of our training method under the image reconstruction depth-learning paradigm, with contributions highlighted. 
     Here we use a video sequence with known egomotion as the signal source. 
     $\tX$ stands for images, $\tD$ for depth map, $\tP$ for point cloud, $\tR$ is projected (registered) point cloud. 
     Superscript $t$ denotes the image sequence number.}
\label{fig:overall}
\end{figure*}

To facilitate this process, the projection of the scene from one image is inverted to generate 3D point cloud, where each point is associated with a pixel in the original image. 
This point cloud is then transformed (by translation and rotation) to the coordinate system of another viewpoint and finally reprojected to synthesize the image of the second view. 
(See Figure~\ref{fig:overall}.)

The quality of the depth prediction is measured indirectly by comparing the pixels in the reconstructed image to the corresponding pixels in the actual image. 
If the depth prediction is correct, the two sets of pixels should be similar, despite the possible changes in lighting or viewing angle.
Stereo image pairs are usually used as the signal source~\cite{garg,godard}, while a video sequence is also possible~\cite{zhou,mahjourian,wang,guizilini20203d}.
Common evaluation benchmarks for this task include KITTI~\cite{kitti} and Cityscapes~\cite{cityscapes}. 

Several problems can occur during the viewpoint transformation phase of this training paradigm. 
First and foremost, after the viewing angle is rotated some visible points become occluded because two points project onto the same pixel in the second image (cf. Figure~\ref{fig:occlusion}).
For correct image reconstruction, we need to render only the point closest to the image plane; 
any other point is occluded by the closest point and thus invisible from the second camera's viewpoint.
The z-buffer, or depth buffer, is a computer graphics workhorse created to solve precisely this problem.

The occlusion problem has only been indirectly addressed by previous work in learning monocular depth estimation.
Godard~\etal~\cite{monodepth2} propose a minimum reprojection loss, and Gordon~\etal~\cite{gordon2019depth} propose a minimum depth consistency filter to simulate a z-buffer.
However, the former solution requires multiple source images and is not guaranteed to eliminate the error, while the latter is still only a heuristic method that can incur both false positives and false negatives.
Using a z-buffer to solve the occlusion problem is conceptually easy but challenging to implement, in that it may be non-trivial to parallelize.
Without caution, the resulting algorithm might become a bottleneck to training.
We propose an efficient, parallel approach (implemented in PyTorch) to address the occlusion problem (see Figure~\ref{fig:demo}). 
Our method is exact (\ie, does not rely on any heuristics) and can be easily incorporated into any existing self-supervised learning framework that relies on reprojection.

Another issue that might be fatal for networks trained from scratch is transformed points with negative depth. 
Having negative depth means that the points are behind the image plane and thus should not be associated with any pixels in the second image. 
Therefore, they are sure to be excluded from the image reconstruction loss. 
However, during the initial phase of our training, when the depth prediction is not very accurate, we found that often many points would get abnormally shallow depth and hence negative depth (\ie, behind the image plane) after transformation. 
Excluding all of those points will make learning inefficient. 
Most of these points should not in fact have negative depth if they can still be projected to a pixel within the second image boundary. 
We therefore introduce a simple loss function to penalize such negative depths.

Experimental evaluation (Section~\ref{sub:results}) demonstrates the importance of these techniques using ablation studies, as well as comparison with previous methods.
We also show that the best timing for inserting the z-buffer into the training process is not at the beginning, but after an initial training phase. 

In summary, our primary contributions include:

\begin{itemize}
\item we confirm that proper z-buffering improves depth estimates for reprojection-based training methods,
\item we provide an efficient z-buffering algorithm compatible with differentiable deep learning systems, and 
\item we propose a loss penalizing erroneously shallow (\ie, negative) depth, which also improves performance.
\end{itemize}

In Section~\ref{sec:related-works} we situate this work in the context of related works, while Section~\ref{sec:method} details our approach. 
Section~\ref{sec:experiments} provides an experimental evaluation of the proposed methods.


\section{Related Works}\label{sec:related-works}

\subsection{Supervised Monocular Depth Estimation}\label{sub:related-works-supervised}

Historically, estimating depth from a single image involved using hand-crafted features or geometric constraints~\cite{survey}. 
The accuracy of those methods often depended on hand-crafted features, which can yield unsatisfactory performance on various scenes if the chosen features are not optimal.
Subsequently, research has shifted toward learning-based methods. 

Depth estimation using deep convolutional neural networks (DCNNs) has followed on the success of DCNNs in many computer vision tasks such as image classification~\cite{resnet} and image segmentation~\cite{ronneberger}. 
Eigen~\etal~\cite{eigen} proposed a multi-scale architecture with both a local and global network to generate a refined, high-resolution depth map from lower resolution depth maps. 
Their work treats the depth estimation task as a regression problem by minimizing the total pixel-wise error between the network output and the ground-truth depth. 
Laina~\etal~\cite{laina} introduced a fully convolutional network with a residual-connected encoder and upsampling block to increase spatial resolution of the predicted depth map. 
Liu~\etal~\cite{liu2017learning} presented a spatial propagation network for predicting an affinity matrix; 
Chen~\etal~\cite{cheng2018depth} subsequently applied the network to the depth estimation task, which increases the resolution of sparse ground truth depth maps. 

To incorporate multi-scale information and high-resolution depth maps, many works adopt a U-net-like architecture, with skip connections between encoder and decoder~\cite{ronneberger}.
With the general success of the encoder-decoder architecture, many improvements have been made, such as substituting the encoder with pre-trained networks~\cite{alhashim2018high} and 
enforcing a planar constraint on local patches of the predicted depth output~\cite{bts}. 

\subsection{Unsupervised Monocular Depth Estimation}

While supervised methods usually offer depth estimation with higher accuracy, the ground truth depth maps used in supervised depth estimation are often subject to spatial inaccuracy due to calibration error and instrument error. 
In addition, ground truth depth measurements are usually quite sparse, arising from 3D LiDARs. 
Therefore, recent research has introduced depth estimation network architectures that learn without depth as an explicit training signal.
In this case, the models use image reconstruction as the supervisory signal during the training stage, as described in Section~\ref{sec:intro}.
The model training process usually takes a temporal series of monocular images and/or pairs of stereo images as input; 
learning proceeds by enforcing the consistency between an observed image and an image from an alternate viewpoint as reconstructed from the depth prediction (as in Figure~\ref{fig:overall}).

Garg~\etal~\cite{garg} proposed the first such unsupervised depth estimation network to achieve performance comparable to supervised networks. 
It uses an encoder-decoder structure to produce a depth for the left image of the stereo image pair and subsequently utilizes the estimated depth of left image, inter-view displacement and the right image to synthesize the left viewpoint. Then, the reconstruction error between the left image and the synthesized left image is used as the training signal for the network.
Zhou~\etal~\cite{zhou} trained a depth estimation network along with a separate pose network and use a monocular video sequence with egomotion instead of a stereo image pair.
To indirectly address the occlusion problem, Zhou~\etal~added an additional explainability mask that excludes pixels for which the network has low confidence in its depth predictions.

Broadening the scope of what should be considered for self-consistency, Godard~\etal~\cite{godard} proposed a network that enforces the left-to-right and right-to-left consistency with smoothness, reconstruction, and left-right disparity consistency losses. 
Mahjourian~\etal~\cite{mahjourian} presented a network architecture using a video sequence as input and enforcing the 3D geometry consistency within an image sequence by utilizing the estimated egomotion.
Godard~\etal~\cite{monodepth2} improved their previous architecture~\cite{godard} by using both the stereo image pair and temporal video sequences as supervisory training signals. 

As mentioned above, most top-performing unsupervised methods learn to predict depth by enforcing the consistency between actual observations and images reconstructed from depth estimates.

Other novel approaches include the work by Guizilini~\etal~\cite{guizilini20203d}, which introduced a velocity supervision loss to solve the scale ambiguity in self-supervised depth estimation networks. 
Luo~\etal~\cite{luo2018} incorporated a GAN architecture, training the model to synthesize the alternate viewpoint's image, and directly calculating depth from the original image and the synthesized image.

\subsection{Occlusion Handling in Reprojection}\label{sub:related-occlusion}

When the loss functions used for training involve photometric agreement between a reconstruction and an actual image, accurate renderings will be important. 
As indicated previously, most early works used all points in the loss function, even if they were occluded in one of the image pairs~\cite{godard,mahjourian}.

More recent work has taken the occlusion into account with attempts to mitigate the issue.
Monodepth2 by Godard~\etal~\cite{monodepth2} incorporates a minimum reprojection loss that requires multiple source images.
This method assumes that a point occluded in one of the source images might be still visible in others.
The images where occlusion happened will likely render higher reconstruction loss due to the erroneously matched pixels. 
Thus, for each pixel in the original image, Monodepth2 uses only the smallest reconstruction loss from several source images, where occlusion is least likely to have happened.
However, this is not applicable when we only have a pair of images (thus only one source image available), as in stereo methods. 
More importantly, lower loss does not guarantee that the chosen source pixel is visible.
PackNet-SfM by Guizilini~\etal~\cite{guizilini20203d} also  adopted this approach. 

Gordon \etal~\cite{gordon2019depth} propose a minimum depth consistency filter to simulate a z-buffer.
Using the predicted depth from frame A as a reference, when points from frame B are transformed to coordinate frame A, any point that is ``behind'' (having greater depth than) the \emph{predicted depth} will be excluded from the loss calculation. 
This heuristic method to eliminate occluded points is sensitive to the model's prediction accuracy, which itself is the learning goal. 

Our work incorporates a more effective method to handle pixel occlusions in the image reconstruction process.
To address the occlusion problem directly, we propose using a z-buffer. 
This is perhaps conceptually easy but challenging to implement, in that it is not fully ``embarrassingly parallel'' where multiple points compete for rendering to the same raster location.
However, efficiency is not the only concern.
Because the z-buffer contents depend on the very depth prediction process being learned, it is likewise potentially sensitive to model accuracy.
At the early stages of training, most depth predictions are inaccurate, and thus points may appear occluded due to the prediction error rather than the true scene geometry. 
Excluding these points from the loss calculation might hamper learning.
On the other hand, if we utilize the z-buffer too late, the model might have already mis-fit, so that the z-buffer might not help or else would require additional training epochs.
Our experiments will demonstrate there is indeed an optimal middle-ground for including the z-buffer.

In theory, any unsupervised method using image reprojection should benefit from incorporating our proposed method.
Our experiments confirm that the performance of an existing model can indeed be further improved by incorporating the method.

\subsection{Z-Buffer Algorithms}\label{sub:related-z-buffer}

With the z-buffer's longstanding history in computer graphics~\cite{zbuffer}, this work is not the first to demand an efficient algorithm.
For graphics rendering, the computation is nearly universally implemented almost entirely in hardware.
However, to be used within a deep learning pipeline, the computation must also produce derivatives, which are not readily available in existing hardware and APIs.

A serial algorithm would calculate a projected pixel coordinate for each 3D point.
If another point is already stored at that coordinate, the algorithm compares their depths and retains only the point with smaller depth value.
In a highly parallelized deep learning pipeline, the speed bottleneck created by such a serial algorithm proves unacceptable.

Work on parallelizing the z-buffer algorithm is nearly as old as the algorithm itself~\cite{fuchs77distributing}. 
We provide a few highlights here.
Forty years ago, Parke~\cite{parke1980simulation} investigated three algorithms that distribute the ``scan conversion'' task (\ie, perspective projection) among parallel processors with a distributed z-buffer.

Li and Miguet~\cite{li1991z} proposed two complementary parallel z-buffer algorithms for a distributed memory transputer. 
They note that a na\"{i}ve approach would distribute \emph{all} scene points to each processor, while dividing the rendering task (portions of the reconstructed image) among processors.
To handle large point clouds, their two algorithms examine either the case when scene elements are statically mapped to a processor, with each process potentially contributing to any rendered pixel, or else the case when scene elements are ``dynamically redistributed'' among a ring of processors.
The first approach requires a (tree-shaped) reduction familiar to modern GPU programmers, with a conditional in the merge that checks for the lower z value.
Motivated by memory limitations, the second approach distributes work and memory demands by dividing the image into regions and rotating subsets of points through each region for processing and merging.


Renaud~\cite{renaud1997fast} similarly concentrates on the difficulty of distributing scene elements to the image regions onto which they project and subsequently balancing the load among SIMD processing elements. 
Like Li and Miguet, Shen~\etal~\cite{shen98efficient} also focus on efficiently merging depths from different objects for the same pixels on a PRAM machine.

Rather than focus on the merge step (which introduces conditionals that lower SIMD GPU throughput), our approach admits a race condition among scene elements mapped to the SIMD processors. 
It then iteratively identifies points occluded points that incorrectly won the race.
The method, described in Section~\ref{sub:zbuffer}, is implementable in modern GPU-based machine-learning libraries, allowing its use in gradient descent frameworks.

\section{Method} \label{sec:method}

In this section, we define the background terms and notation before describing our overarching training method. 
We then detail solutions to several issues that occur during the point cloud transformation phase of the self-supervised depth learning paradigm, as introduced in Sections~\ref{sec:intro} and~\ref{sec:related-works}.

\subsection{Definitions}\label{sub:definitions}

In this paper, the notation font for tensors is $\tA$, for matrices is $\mA$, and for scalars is $A$. 
The variables used in our training method include $\tX\in \mathbb{R}^{3\times h \times w}$ for RGB images, $\tD\in \mathbb{R}^{1\times h \times w}$ for depth images, $\tP\in \mathbb{R}^{3\times n}$ for point clouds, $\tR\in \mathbb{R}^{3\times h \times w}$ for point cloud registration (explained below), $\mE$ for the relative position matrix between cameras, and $\mM_{\textrm{proj}}$ for the projection matrix (i.e., the camera intrinsic matrix).

Image sequence number is indicated using superscript (\eg, $\tX^t$, $\tP^{t+1}$), and pixel location is indicated using subscript (\eg, $\tX_{i,j}$, $\tR_{i,j}$).

The point cloud registration $\tR$ is a tensor such that $\tR_{i,j}$ is the $(x,y,z)$ coordinates of the point in 3D space corresponding to the pixel at location $(i,j)$ in image $\tX$.

The $\ell_1$ loss (sum of absolute values) is denoted $\Vert\cdot\Vert_1$.

\subsection{Basic Model}\label{sub:model}

Our overarching training method uses image reconstruction (Figure~\ref{fig:overall}), combined with point cloud matching. 
We first take two images $\tX^t$ and $\tX^{t+1}$, with their relative position $\mE^t$ known.
For stereo data, the images are ``left'' and ``right'' pairs with a temporally invariant relative position $\mE$, but the $t$, $t+1$ sequence applies more generally to egomotion sequences as well.
We train a model to predict depth $\tD^t$ and $\tD^{t+1}$ for each of the images and use the known camera intrinsic matrix to inverse project the depths to point clouds $\tP^t$ and $\tP^{t+1}$. 
We then transform the point clouds to each other's relative position. Using the matrix $\mE^t$ we have
\begin{equation}\label{eq:model-point-cloud-recon-left}
    \hat{\tP}^t \triangleq \mE^t\tP^{t+1}.
\end{equation}
($\mE^t$ is invertible, allowing for the reverse process.)
Finally, we project these transformed point clouds onto the image plane, get the resulting pixel coordinates for each point, and sample the pixel color from the other image (e.g., $\tX^{t+1}$) to reconstruct the original image from the other viewpoint (e.g., $\hat{\tX}^{t}$).

Although the following loss functions are all written from one image to another, both directions are included in the total loss during training. 

A simple point cloud matching loss~\cite{godard}
\begin{equation}\label{eq:loss_p}
    L_{\textrm{point}}=\Vert\hat{\tR}^t-\tR^{t}\Vert_1
\end{equation}
ensures the consistency of point clouds predicted for continuous image frames, where $\tR^t$ is the original registration tensor simply reshaped from $\tP^t$.
That is, $\hat{\tR}^t$ stores the points from $\hat{\tP}^t$ that are registered to the pixel locations in $\tX^t$ using the z-buffering algorithm introduced in Section~\ref{sub:zbuffer}. 

An image reconstruction loss~\cite{zhou,garg}
\begin{equation}\label{eq:loss_i}
    L_{\textrm{image}}=\Vert\hat{\tX}^t-\tX^{t}\Vert_1
\end{equation}
ensures the similarity between the original image and the reconstructed image from the transformed point cloud. 
We also include the Structured Similarity (SSIM) Loss used in Godard~\etal~\cite{godard} and Mahjourian~\etal~\cite{mahjourian}:
\begin{equation}\label{eq:loss_ssim}
    L_{\textrm{SSIM}}=\sum\limits_p 1- \textrm{SSIM}(\hat{\tX}_p^t,\tX_p^{t}),
\end{equation}
where $\tX_p$ here represents a $3\times 3$ image patch. 
Our overall loss function is 
\begin{equation}\label{eq:total}
    L_{\textrm{total}}=\lambda_1L_{\textrm{point}}+\lambda_2L_{\textrm{image}}+\lambda_3L_{\textrm{SSIM}}+\lambda_4L_{\textrm{nd}},
\end{equation}
where $L_{\textrm{nd}}$ denotes the ``negative depth loss," detailed below in Section~\ref{sub:negdep}. 
Section~\ref{sub:details} lists the specific relative weights $\lambda_i$ and other hyperparameter details.

\subsection{Negative In-frame Depth}\label{sub:negdep}

\begin{figure}
\begin{center}
{\includegraphics[width=0.9\linewidth]{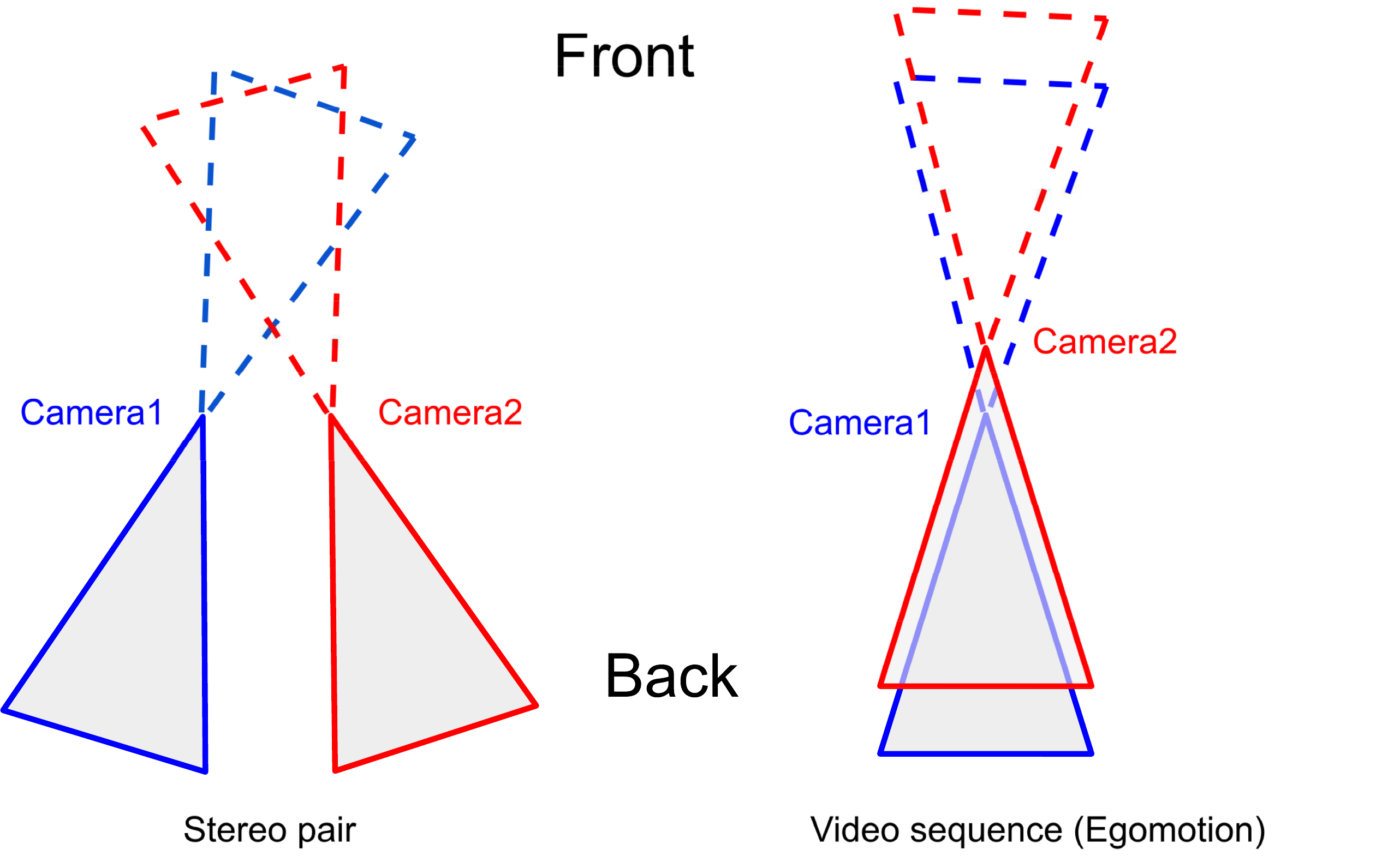}}
\end{center}
   \caption{Improbability of negative depth. 
   It is highly unlikely for points visible to camera 1 to be in frame but have negative depth with respect to camera 2 (i.e., to appear in the grey area of camera 2). 
   With egomotion, such points need to be squeezed between the two cameras (inside the little shaded diamond area in the right figure), which is extremely unlikely to happen in data sets like KITTI~\cite{kitti}.}
\label{fig:negdep}
\end{figure}

After a point cloud has been transformed to another viewpoint and reprojected, some points might fall behind the image plane, resulting in a negative depth.
Stereo pairs or video sequences have relatively close viewpoints.
Thus it is highly unlikely that a point originally projecting to the center of the first image will become invisible in the second image (see Figure \ref{fig:negdep}). 
When negative depth predictions occur, they signify that the model is producing abnormally or inappropriately shallow depth values. 

Let $\mathcal{N}$ be set of points with negative depths that remain within the image boundary after transformation and projection:    
\begin{equation}\label{eq:negdepth}
\mathcal{N} \triangleq \left\{ \left(i,j\right) | \left( d_{i,j} < 0 \right) \wedge \left( 0\leq i<w \right) \wedge \left( 0 \leq j<h\right)\right\},
\end{equation}
where $\left(i,j\right)$ is the image coordinate of a point, and $d_{i,j}$ is the depth of the point. Our loss term penalizes such points:
\begin{equation}\label{eq:loss_nd}
    L_{\text{nd}} = \sum\limits_{\left(i,j\right) \in \mathcal{N}}\left|d_{i,j}\right|.
\end{equation}

We observe that it is not meaningful to include the in-frame, negative-depth points of $\mathcal{N}$ in the loss calculations for $L_{\text{point}}$ (\eqref{eq:loss_p}), $L_{\text{image}}$ (\eqref{eq:loss_i}), and $L_{\text{SSIM}}$ (\eqref{eq:loss_ssim}).
We are not aware of prior work that filters these points from the calculations (and thus from learning).
In our experiments below, when accounting for negative in-frame depth we also exclude points in $\mathcal{N}$ from all losses except $L_{\text{nd}}$,~\eqref{eq:loss_nd}. We found these conditions to improve performance.

We also found that this loss is particularly helpful during the initial phase of the unsupervised training, especially when the encoder is not pre-trained. 
If the network tends to predict erroneously shallow depth at the beginning, this loss can help push the depth value toward the correct range. 
If we do not penalize this erroneous negative depth, but instead only mask them out, then most all the points are discarded and the network is unable to learn.

\subsection{Efficient Z-buffering}\label{sub:zbuffer}

\begin{figure}
\begin{center}
\includegraphics[width=0.9\linewidth]{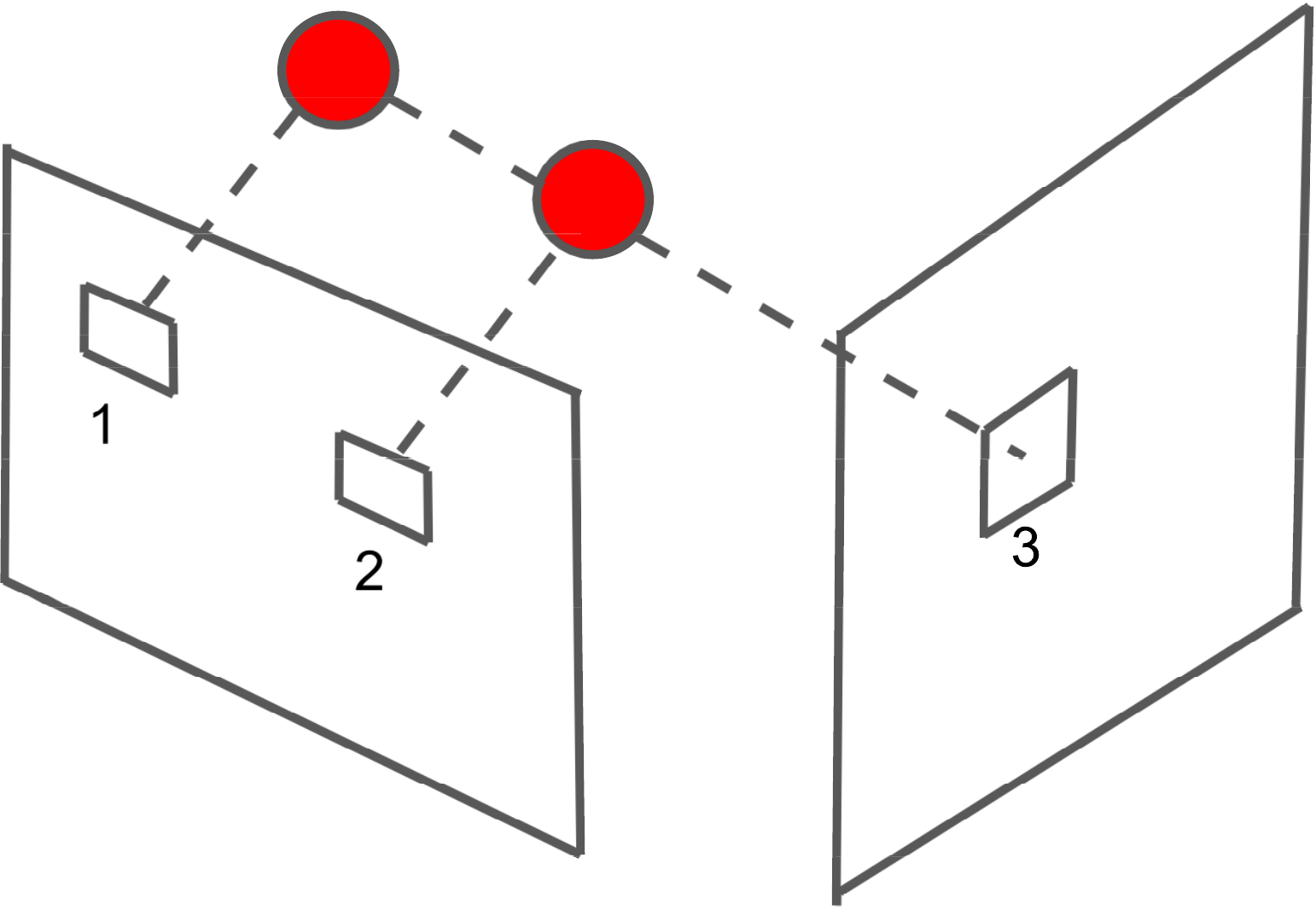}
\end{center}
   \caption{The occlusion problem: two points in space projecting to different pixels in the original image might be projected to the same pixel on another image.
     Pixel 1 should not be used to match pixel 3 in the image reconstruction loss.}
\label{fig:occlusion}
\end{figure}

Both the $L_\textrm{point}$ loss and the $L_\textrm{image}$ loss of \eqref{eq:total} require the correspondence of pixels between two images. 
As Figure~\ref{fig:occlusion} shows, an occluded point in one frame should not correspond to any point in the other frame.
Occlusion happens after the point cloud transformation phase because some points inevitably ``overlap" with each other by projecting onto the same pixel in the second image.
In these cases, we need to choose the point closer to the image plane, as it is the point associated with the visible pixel. 

\begin{algorithm}
\SetAlgoLined
\SetKwInOut{Input}{Input}
\SetKwInOut{Output}{Output}
\Input{$\tD$: $1\times N$, reprojected point depths \\
$\tK$: $1\times N$, point raster absolute indices}
\Output{$\tV$, indices for points that should be included in the loss.}
 initialize empty $\tZ$, same size as $\tD$ \tcp{Z- buffer}
 $\tD_\textrm{orig}\leftarrow \tD$ \\
 $\tK_\textrm{orig}\leftarrow \tK$ \\
 \Repeat{$\left|\tU\right| = 0$}{
 \tcp{ Parallel write; race condition}
  $\tZ[\tK[p]] \leftarrow \tD[p]$, for all $0\leq p < \left|\tD\right|$ \\
  create $\tT$, same size as $\tD$ \\
  \tcp{Parallel read of assigned depths}
  $\tT[p] \leftarrow \tZ[\tK[p]]$, for all $0\leq p< \left|\tD\right|$ \\
  \tcp{Find occluded points in Z-buffer}
  $\tU \leftarrow \left\{ p \mid \tD[j] < \tT[p] \right\}$ \tcp{Parallel compare} 
  \If{$\left|\tU\right| \not= 0$}{
     $\tD\leftarrow \tD[\tU]$ \tcp{"Contract" to nearer}
     $\tK\leftarrow \tK[\tU]$ \tcp{points NOT in Z-buffer}
  }
 }
 create $\tT$, same size as $\tD_\textrm{orig}$ \\
 $\tT[p] \leftarrow \tZ[\tK_\textrm{orig}[p]]$ for all $0\leq p< N$ \\
 $\tV \leftarrow \left\{ p \mid \tD_\textrm{orig}[p] = \tT[p] \right\}$ \tcp{Visible points}
 \caption{Parallel z-buffering algorithm.}\label{al:zbuffer}
\end{algorithm}

To fundamentally solve the issue, we need to identify the point closer to the image plane when occlusion occurs. 
Our method is inspired by the z-buffer (or depth buffer) concept from computer graphics, which can achieve efficient processing for all points using parallel computing (cf. Section~\ref{sub:related-z-buffer}). 
We explain Algorithm~\ref{al:zbuffer} below. By implementing it in PyTorch, we easily access its differentiable, parallelized operators. 

Given a set of points from frame A, we transform them into coordinate frame B. 
The transformed depth $\tD$ is of shape $1\times h \times w$. 
We vectorize it into $1\times N$, and calculate the corresponding $(i,j)$ pixel coordinates $\tC\in \mathbb{R}^{2\times N}$ on image B for all the points using the camera intrinsic matrix. 
We employ the ``principled mask"~\cite{mahjourian} to discard the points that fall out of the image frame: masking out the points where the $i$ ($j$, resp.) are larger than $h-1$ ($w-1$, resp.) or smaller than $0$. 
We also discard points in $\mathcal{N}$, \eqref{eq:negdepth}, which are within the image frame but have negative depth values, as discussed in Section \ref{sub:negdep}.

For each point that remains, we calculate the ``absolute indices" $\tK$ ($1\times N$) using the coordinate values from $\tC$, 
\begin{equation}\label{eq:abs-index}
k \triangleq j\times w + i.
\end{equation} 
Thus, \textbf{points projecting to the same pixel location have the same absolute index.}

Let $\tZ$ of shape $1\times N$ be a blank tensor to store the correct (visible) depth at each pixel location. 
Now we assign the elements of $\tD$ into $\tZ$ using absolute indices $\tK$ (line 5 in Algorithm~\ref{al:zbuffer}).
When multiple points project to the same pixel, this operation results in a race condition; we will not know whether the assigned point is visible.
This approach stands somewhat in contrast to the prior work reviewed in Section~\ref{sub:related-z-buffer}, which carefully folds competing points through conditional assignments.
In our algorithm, to compare the values stored in the z-buffer $\tZ$ with the original depths $\tD$, we fetch back depth values from $\tZ$ to a temporary tensor $\tT$ (line 7), inverting the assignment.
Then we simply compare the values in $\tD$ and $\tT$ to retain only points with a smaller depth value than stored in $\tT$ (line 8). 
Having selected only those points from $\tD$ and $\tK$, we repeat the process (overwriting occluded points in the z-buffer) until there is no difference between $\tD$ and $\tT$.

During our experiments, this process iterates three or four times before all the correct depth values are stored, with only a few points left in the third or fourth rounds. 

Finally, we use the correct $\tZ$ tensor and compare it with the original $\tD$ tensor. 
The positions where the two values are the same indicate the points that should be included in our losses.

Note the length of $\tU$ becomes strictly smaller with each iteration. 
With a finite number of points, the algorithm is guaranteed to terminate. 
With this method, we can quickly process all points in parallel, with the iteration limit being the number of maximum number of occluding points along a ray.

Because all the operations in the algorithm can be implemented with PyTorch operators, the GPU parallelization is straightforward.
Moreover,the implementation gains the necessary benefit of providing the derivatives for the learning pipeline. 
For example, the assignment of line 5 in Algorithm~\ref{al:zbuffer} is implemented in PyTorch using \texttt{Tensor.index\_copy} (or Tensorflow \texttt{tf.scatter\_nd}), while the read out of lines 7 and 15 use \texttt{Tensor.index\_select} (or \texttt{tf.gather\_nd}). 
Lines 8 and 16 also employ SIMD-parallel comparators followed by select/gather.

\section{Experiments}\label{sec:experiments}

This section describes the results of our model on standard data sets, comparing against other recent state of the art works and demonstrating the performance improvements of our approach.

\begin{table*}

\caption{Performance on KITTI Eigen test split. 
  Supervision signal: D - depth, V - video sequence (w/o ground-truth egomotion), S - stereo.
  $\downarrow$ means lower is better, $\uparrow$ means higher is better. Best overall in \textbf{bold}. Best among S in \bestste{blue}. Best among V in \bestvid{red}.
  }
\begin{center}
\begin{tabular}{c|c|cccc|ccc}
Method & Signal & Abs Rel$\downarrow$ & Sq Rel$\downarrow$ & RMSE$\downarrow$ & RMSE log$\downarrow$ & $\delta < 1.25\uparrow$ & $\delta < 1.25^2\uparrow$ & $\delta < 1.25^3\uparrow$\\\hline
Eigen \etal \cite{eigen} & D & 0.203 & 1.548 & 6.307 & 0.282 & 0.702 & 0.890 & 0.958 \\
Zhou \etal \cite{zhou} * & V & 0.198 & 1.836 & 6.565 & 0.275 & 0.718 & 0.901 & 0.960\\
Garg \etal \cite{garg} $\dagger$  & S &0.169& 1.080& 5.104& 0.273& 0.740 &0.904 &0.962\\
Mahjourian \etal \cite{mahjourian} *& V &0.159& 1.231& 5.912& 0.243& 0.784& 0.923 &0.970\\
Wang \etal \cite{wang} *& V & 0.148 &1.187 &5.496 & 0.226 &0.812& 0.938 & 0.975\\
Godard \etal \cite{godard} * & S &0.114 & 0.898 & 4.935&0.206 & 0.861 & 0.949 & 0.976\\
Gordon \etal \cite{gordon2019depth} & V & 0.129 & 0.982 & 5.230 & - & - & - & - \\
Godard \etal \cite{monodepth2} & S & 0.107 & 0.849 & 4.764 & \bestste{0.201} &\bestste{0.874} & \bestste{0.953} & \bestste{0.977}\\
Godard \etal \cite{monodepth2} & S + V & \textbf{0.106} & 0.806 & 4.630 & 0.193 & 0.876 & 0.958 & 0.980\\
Guizilini \etal \cite{guizilini20203d} & V & \bestvid{0.107} & \bestvid{0.802} & \bestvid{\textbf{4.538}} & \bestvid{\textbf{0.186}} & \bestvid{\textbf{0.889}} & \bestvid{\textbf{0.962}} & \bestvid{\textbf{0.981}} \\
This work & S & \bestste{\textbf{0.106}} &  \bestste{\textbf{0.743}} & \bestste{4.707} &      \bestste{0.201} & 0.864 & 0.949 & \bestste{0.977} \\\hline
\end{tabular}
\end{center}
\begin{footnotesize} 
* method is trained on Cityscapes data set \cite{cityscapes} and fine-tuned on KITTI.

$\dagger$ method is tested with depth capped at 50m, otherwise 80m. 
\end{footnotesize}
\label{tb:eval}
\end{table*}

\begin{table*}

\caption{Ablation studies conducted on KITTI Eigen test split.
$\downarrow$ means lower is better, $\uparrow$ means higher is better.}

\begin{center}
\begin{tabular}{c|cc|cccc|ccc}
Negative   & Occlusion &  Insertion &  \multirow{2}{*}{Abs Rel$\downarrow$} & \multirow{2}{*}{Sq Rel$\downarrow$} & \multirow{2}{*}{RMSE$\downarrow$} & \multirow{2}{*}{RMSE log$\downarrow$} & \multirow{2}{*}{$\delta < 1.25\uparrow$} & \multirow{2}{*}{$\delta < 1.25^2\uparrow$} & \multirow{2}{*}{$\delta < 1.25^3\uparrow$}\\
Depth Loss & Handling  &    Epoch & & & & & &  \\\hline
 - & none & - & 0.108&     0.776&      4.881&      0.210&      0.854&      0.946&      0.974\\
$\checkmark$ & none & - & 0.109 & 0.764 & 4.790&      0.207&       0.859&     \textbf{0.949}&      0.975 \\
\hdashline
$\checkmark$ & z-buffer & 1 & 0.108&     0.764&      4.857&      0.207&         0.855&      0.948&      0.976 \\
$\checkmark$ & z-buffer & 6 & 0.109&     0.775&      4.847&      0.211&      0.851&      0.943&      0.974 \\
$\checkmark$ & z-buffer & 11 & \textbf{0.106} &  \textbf{0.743} & \textbf{4.707} &      \textbf{0.201} & \textbf{0.864} & \textbf{0.949} & \textbf{0.977} \\
$\checkmark$ & z-buffer & 16 & 0.109&     0.770&      4.898&      0.210&        0.854&      0.946&      0.974 \\
\hdashline
- & z-buffer & 11 & 0.109&    0.752&      4.808&      0.205&          0.858&     \textbf{0.949}&      0.976\\
$\checkmark$ & heuristic~\cite{gordon2019depth} & 11 & 0.115&     0.860&      5.111&      0.223&         0.846&      0.939&      0.970\\\hline
\end{tabular}
\end{center}
\label{tb:ablation}
\end{table*}

\subsection{Data Set}

For all training and evaluation we use the KITTI data set~\cite{kitti} with the Eigen~\cite{eigen} training split. 

For validation, we sample a random subset of size 500 from the Eigen validation split before training starts. 
After every 200 steps, we evaluate the model on the validation set, preserving the best model checkpoint so far (using the $\delta<1.25$ metric).

Finally, for evaluation, we use the standard Eigen~\cite{eigen} test split, which contains 697 samples in total. 
The ground-truth depth is calculated from the laser point clouds provided by KITTI. 
The depth is capped at 80 meters. 
We use the crop standard of Garg~\etal~\cite{garg}, cropping away the upper half of the image and some boundary parts to avoid comparing with inaccurate ground-truth.
We evaluate on the best checkpoint identified during training using the validation set. 

\subsection{Model Implementation and Training Details}\label{sub:details}

For the depth prediction model, we use the ``big-to-small'' multi-scale local planar guidance architecture of Lee~\etal~\cite{bts}; the encoder is ResNet-50~\cite{resnet}, pre-trained on ImageNet~\cite{imagenet}, with all weights adjusted during training.

We train for 20 epochs with a batch size of 3 using the Adam optimizer ($\beta_1=0.9$, $\beta_2=0.999$) and a learning rate of $1\textsc{e--}05$. 

The images in the KITTI data set vary somewhat in size. 
To compensate, we crop all training images to $352\times 1216$, aligning the upper-left corner. 

We augment the image data by applying small random variations to the colors of the input images given to the depth prediction network.
However, we sample unmodified pixel colors from the original images during image reconstruction to avoid spurious color mismatches.

The hyperparameters of $L_\textrm{SSIM}$ in~\eqref{eq:loss_ssim} are as given in Mahjourian~\etal~\cite{mahjourian}.
For our overall loss function in \eqref{eq:total}, we set $\lambda_1=0.005$, $\lambda_2=10$, $\lambda_3=2$ and $\lambda_4=2$, which puts each loss term on roughly the same scale and acknowledges the relatively low importance of point cloud matching after the initial training.

\subsection{Results}\label{sub:results}

Table~\ref{tb:eval}  lists the evaluation results on the standard Eigen~\cite{eigen} test split compared with several previous methods. 
All the methods listed are \emph{not} fine-tuned on any ground-truth depth data.
This table is not meant to demonstrate our approach to be superior over all---although it does boost performance on some metrics---but instead to demonstrate its comparability with other similar unsupervised methods on the standard metrics, particularly when using the same training signals.
In addition, it gives context for the relative scale of improvements.

Rather, the ablation experiments of Table~\ref{tb:ablation} demonstrate the contributions of our methods with respect to a fixed network architecture.
The first row establishes a baseline, using the network trained without the negative depth loss (\ie, $\lambda_4=0$) and no occlusion-handling method, explicit or implicit.
Importantly, when the negative depth loss is excluded, we also include in-frame negative-depth points from $\mathcal{N}$ in the other losses, as in prior work.
The second row demonstrates that the incorporating the negative depth loss (and excluding $\mathcal{N}$ from other losses) yields an improvement across all but one metric.
The next group of several rows incorporate the z-buffer for occlusion handling at different points in the training process (\ie, beginning, 25\%, 50\%, and 75\% of the way through).
Having found an optimal time to insert the z-buffer (epoch 11 of 20, 50\% of the way through the training), the last two rows test the relative contributions of the negative depth loss and the alternative occlusion-handling heuristic proposed by Gordon~\etal~\cite{gordon2019depth}. 

These results demonstrate that the z-buffer indeed improves the depth prediction results when inserted at the right stage of training.
Inserting z-buffer either too early or too late harms the performance, which is not surprising. 

At the early stages of training, most depth predictions are inaccurate, and thus the points appear occluded due to the prediction error rather than the true scene geometry. 
Excluding the points from the loss calculation hampers learning.
On the other hand, if we utilize the z-buffer too late, then truly occluded points might inappropriately factor into the loss function and confuse the model.

Results also show that our use of a z-buffer outperforms the consistency-driven heuristic method of Gordon~\etal~\cite{gordon2019depth}, even when inserted to the training pipeline with the same timing. 
Their method excludes every transformed point that is ``behind" the predicted depth of the target image (cf.~Section~\ref{sub:related-occlusion}). 
Experimental results imply that even when the depth predictions are roughly reliable (around epoch 11), we should not expect points from original frame and target frame to agree precisely.
Some tiny amount of error can make a valid rotated point fall behind the target depth image, masking it from the loss calculation.
With so many extra points excluded from training, the model does not learn as well.

During our experiments, we found that excluding points with negative depth from the loss calculation and using a negative-depth loss is crucial for training shallow networks from scratch.
Without the exclusion, the model will quickly be led astray by the erroneous depth information. 
For pre-trained deep networks, including such points and turning off the negative-depth loss is not fatal, but still harms performance (as shown in Table~\ref{tb:ablation}).

\section{Conclusions}

Our work proposes solutions to important issues in the point transformation phase of the image reconstruction training paradigm for self-supervised methods for monocular depth estimation.
We experiment on stereo pairs, demonstrating the effectiveness of our methods on the model's performance compared to past literature and through ablation studies.

Our results demonstrate that the parallelized z-buffering algorithm rectifies inconsistencies in the loss functions involving reprojections, allowing for improved learning and test performance.
Moreover, our algorithm can be easily incorporated with other self-supervised approaches to monocular depth prediction that use reprojection and associated metrics.

In addition, we demonstrate that penalizing in-frame points with negative depth, an unlikely situation, can also improve model performance.

Because they are rooted in a general approach rather than a particular network architecture, these procedural changes potentially offer benefits to a wide variety of monocular depth prediction methods.

\bibliographystyle{IEEEtran}
\bibliography{main}

\begin{thebibliography}{10}
\providecommand{\url}[1]{#1}
\csname url@samestyle\endcsname
\providecommand{\newblock}{\relax}
\providecommand{\bibinfo}[2]{#2}
\providecommand{\BIBentrySTDinterwordspacing}{\spaceskip=0pt\relax}
\providecommand{\BIBentryALTinterwordstretchfactor}{4}
\providecommand{\BIBentryALTinterwordspacing}{\spaceskip=\fontdimen2\font plus
\BIBentryALTinterwordstretchfactor\fontdimen3\font minus
  \fontdimen4\font\relax}
\providecommand{\BIBforeignlanguage}[2]{{%
\expandafter\ifx\csname l@#1\endcsname\relax
\typeout{** WARNING: IEEEtran.bst: No hyphenation pattern has been}%
\typeout{** loaded for the language `#1'. Using the pattern for}%
\typeout{** the default language instead.}%
\else
\language=\csname l@#1\endcsname
\fi
#2}}
\providecommand{\BIBdecl}{\relax}
\BIBdecl

\bibitem{kitti}
A.~Geiger, P.~Lenz, C.~Stiller, and R.~Urtasun, ``Vision meets robotics: The
  {KITTI} dataset,'' \emph{International Journal of Robotics Research (IJRR)},
  vol.~32, pp. 1231--1237, 2013.

\bibitem{garg}
R.~Garg, V.~K. BG, G.~Carneiro, and I.~Reid, ``Unsupervised {CNN} for single
  view depth estimation: Geometry to the rescue,'' in \emph{European Conference
  on Computer Vision}.\hskip 1em plus 0.5em minus 0.4em\relax Springer, 2016,
  pp. 740--756.

\bibitem{godard}
C.~Godard, O.~Mac~Aodha, and G.~J. Brostow, ``Unsupervised monocular depth
  estimation with left-right consistency,'' in \emph{Proceedings of the IEEE
  Conference on Computer Vision and Pattern Recognition}, 2017, pp. 270--279.

\bibitem{zhou}
T.~Zhou, M.~Brown, N.~Snavely, and D.~G. Lowe, ``Unsupervised learning of depth
  and ego-motion from video,'' in \emph{Proceedings of the IEEE Conference on
  Computer Vision and Pattern Recognition}, 2017, pp. 1851--1858.

\bibitem{mahjourian}
R.~Mahjourian, M.~Wicke, and A.~Angelova, ``Unsupervised learning of depth and
  ego-motion from monocular video using 3{D} geometric constraints,'' in
  \emph{Proceedings of the IEEE Conference on Computer Vision and Pattern
  Recognition}, 2018, pp. 5667--5675.

\bibitem{wang}
C.~Wang, J.~Miguel~Buenaposada, R.~Zhu, and S.~Lucey, ``Learning depth from
  monocular videos using direct methods,'' in \emph{Proceedings of the IEEE
  Conference on Computer Vision and Pattern Recognition}, 2018, pp. 2022--2030.

\bibitem{guizilini20203d}
V.~Guizilini, R.~Ambrus, S.~Pillai, A.~Raventos, and A.~Gaidon, ``{3D} packing
  for self-supervised monocular depth estimation,'' in \emph{Proceedings of the
  IEEE/CVF Conference on Computer Vision and Pattern Recognition}, 2020, pp.
  2485--2494.

\bibitem{cityscapes}
M.~Cordts, M.~Omran, S.~Ramos, T.~Rehfeld, M.~Enzweiler, R.~Benenson,
  U.~Franke, S.~Roth, and B.~Schiele, ``The {C}ityscapes dataset for semantic
  urban scene understanding,'' in \emph{Proceedings of the IEEE Conference on
  Computer Vision and Pattern Recognition}, 2016, pp. 3213--3223.

\bibitem{monodepth2}
C.~Godard, O.~Mac~Aodha, M.~Firman, and G.~J. Brostow, ``Digging into
  self-supervised monocular depth estimation,'' in \emph{Proceedings of the
  IEEE International Conference on Computer Vision}, 2019, pp. 3828--3838.

\bibitem{gordon2019depth}
A.~Gordon, H.~Li, R.~Jonschkowski, and A.~Angelova, ``Depth from videos in the
  wild: Unsupervised monocular depth learning from unknown cameras,'' in
  \emph{Proceedings of the IEEE International Conference on Computer Vision},
  2019, pp. 8977--8986.

\bibitem{survey}
A.~Bhoi, ``Monocular depth estimation: A survey,'' \emph{arXiv preprint
  arXiv:1901.09402}, 2019.

\bibitem{resnet}
K.~He, X.~Zhang, S.~Ren, and J.~Sun, ``Deep residual learning for image
  recognition,'' in \emph{Proceedings of the IEEE Conference on Computer Vision
  and Pattern Recognition}, 2016, pp. 770--778.

\bibitem{ronneberger}
O.~Ronneberger, P.~Fischer, and T.~Brox, ``U-net: Convolutional networks for
  biomedical image segmentation,'' in \emph{International Conference on Medical
  image computing and computer-assisted intervention}.\hskip 1em plus 0.5em
  minus 0.4em\relax Springer, 2015, pp. 234--241.

\bibitem{eigen}
D.~Eigen, C.~Puhrsch, and R.~Fergus, ``Depth map prediction from a single image
  using a multi-scale deep network,'' in \emph{Advances in Neural Information
  Processing Systems}, 2014, pp. 2366--2374.

\bibitem{laina}
I.~Laina, C.~Rupprecht, V.~Belagiannis, F.~Tombari, and N.~Navab, ``Deeper
  depth prediction with fully convolutional residual networks,'' in \emph{2016
  Fourth international conference on 3D vision (3DV)}.\hskip 1em plus 0.5em
  minus 0.4em\relax IEEE, 2016, pp. 239--248.

\bibitem{liu2017learning}
S.~Liu, S.~De~Mello, J.~Gu, G.~Zhong, M.-H. Yang, and J.~Kautz, ``Learning
  affinity via spatial propagation networks,'' in \emph{Advances in Neural
  Information Processing Systems}, 2017, pp. 1520--1530.

\bibitem{cheng2018depth}
X.~Cheng, P.~Wang, and R.~Yang, ``Depth estimation via affinity learned with
  convolutional spatial propagation network,'' in \emph{Proceedings of the
  European Conference on Computer Vision (ECCV)}, 2018, pp. 103--119.

\bibitem{alhashim2018high}
I.~Alhashim and P.~Wonka, ``High quality monocular depth estimation via
  transfer learning,'' \emph{arXiv preprint arXiv:1812.11941}, 2018.

\bibitem{bts}
J.~H. Lee, M.-K. Han, D.~W. Ko, and I.~H. Suh, ``From big to small: Multi-scale
  local planar guidance for monocular depth estimation,'' \emph{arXiv preprint
  1907.10326}, 2019.

\bibitem{luo2018}
Y.~Luo, J.~Ren, M.~Lin, J.~Pang, W.~Sun, H.~Li, and L.~Lin, ``Single view
  stereo matching,'' in \emph{Proceedings of the IEEE Conference on Computer
  Vision and Pattern Recognition}, 2018, pp. 155--163.

\bibitem{zbuffer}
W.~Stra{\ss}er, ``Schnelle kurven-und fl{\"a}chendarstellung auf grafischen
  sichtger{\"a}ten,'' Ph.D. dissertation, Technischen Universit\"{a}t Berlin,
  1974.

\bibitem{fuchs77distributing}
H.~Fuchs, ``Distributing a visible surface algorithm over multiple
  processors,'' in \emph{Proceedings of the 1977 ACM Annual Conference}, 1977.

\bibitem{parke1980simulation}
F.~I. Parke, ``Simulation and expected performance analysis of multiple
  processor z-buffer systems,'' in \emph{Proceedings of the 7th Annual
  Conference on Computer Graphics and Interactive Techniques}, ser. SIGGRAPH
  '80, 1980, p. 48–56.

\bibitem{li1991z}
J.-j. Li and S.~Miguet, ``Z-buffer on a transputer-based machine,'' in
  \emph{The Sixth Distributed Memory Computing Conference, 1991.
  Proceedings}.\hskip 1em plus 0.5em minus 0.4em\relax IEEE Computer Society,
  1991, pp. 315--316.

\bibitem{renaud1997fast}
C.~Renaud, ``Fast local and global illuminations through a {SIMD} z-buffer,''
  \emph{International journal of pattern recognition and artificial
  intelligence}, vol.~11, no.~07, pp. 1095--1112, 1997.

\bibitem{shen98efficient}
H.~Shen, J.~You, and D.~J. Evans, ``An efficient parallel algorithm or
  visible-surface detection in 3d graphics display,'' \emph{International
  Journal of Computer Mathematics}, vol.~67, no. 3-4, pp. 359--371, 1998.

\bibitem{imagenet}
J.~{Deng}, W.~{Dong}, R.~{Socher}, L.~{Li}, {Kai Li}, and {Li Fei-Fei},
  ``Imagenet: A large-scale hierarchical image database,'' in \emph{2009 IEEE
  Conference on Computer Vision and Pattern Recognition}, 2009, pp. 248--255.

\end{thebibliography}

\end{document}